\documentclass{article}
\usepackage{spconf}
\usepackage{graphicx}
\usepackage{hyperref}
\usepackage{xcolor}
\usepackage{dblfloatfix}
\usepackage{subcaption}
\begin{document}
\title{Generalized Zero-Shot Learning using Multimodal Variational Auto-Encoder with Semantic Concepts}

\name{Nihar Bendre \qquad Kevin Desai \qquad Peyman Najafirad$^*$}

  \address{  Secure Artificial Intelligent Laboratory and Autonomy (AILA) \\
                            The University of Texas at San Antonio, Texas, USA \\
                          \{nihar.bendre, kevin.desai, peyman.najafirad\}@utsa.edu\\
                          \thanks{$^*$ Corresponding author}}

\maketitle              
\begin{abstract}

With the ever-increasing amount of data, the central challenge in multimodal learning involves limitations of labelled samples
For the task of classification, techniques such as meta-learning, zero-shot learning, and few-shot learning showcase the ability to learn information about novel classes based on prior knowledge .
Recent techniques try to learn a cross-modal mapping between the semantic space and the image space.
However, they tend to ignore the local and global semantic knowledge.
To overcome this problem, we propose a Multimodal Variational Auto-Encoder (M-VAE) which can learn the shared latent space of image features and the semantic space. In our approach we concatenate multimodal data to a single embedding before passing it to the VAE for learning the latent space.
We propose the use of a multi-modal loss during the reconstruction of the feature embedding through the decoder.
Our approach is capable to correlating modalities and exploit the local and global semantic knowledge for novel sample predictions.
Our experimental results using a MLP classifier on four benchmark datasets show that our proposed model outperforms the current state-of-the-art approaches for generalized zero-shot learning.
\end{abstract}


\vspace{-3mm}
\section{Introduction}
\label{sec:intro}
\vspace{-2mm}

Techniques such as few-shot learning, zero-shot learning, and meta-learning are gaining popularity amongst researchers for addressing the inability of deep learning models to learn from limited samples \cite{bendre2020learning}.
The general idea for these techniques is to use knowledge transfer to recognize novel classes by exposing the model to zero or handful of samples during training.
In zero-shot learning (ZSL) \cite{fe2003bayesian}, the train and test data only share the attributes which are annotated class-wise.
The performance of zero-shot learning models is measured by their ability to correctly classify the novel test class.
In real-world settings, it is difficult to have a scenario where the train and the test set are completely disjoint.

\begin{figure}[t!]
\begin{subfigure}{.5\linewidth}
  \centering
  \includegraphics[width=0.9\linewidth]{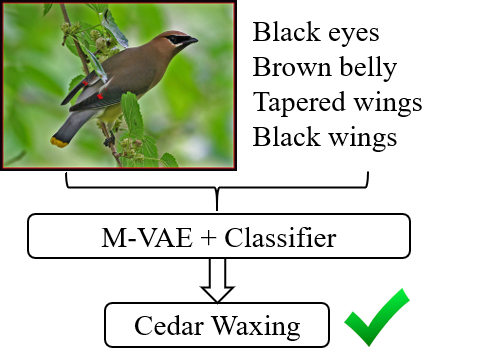}
\end{subfigure}%
\begin{subfigure}{.5\linewidth}
  \centering
  \includegraphics[width=0.9\linewidth]{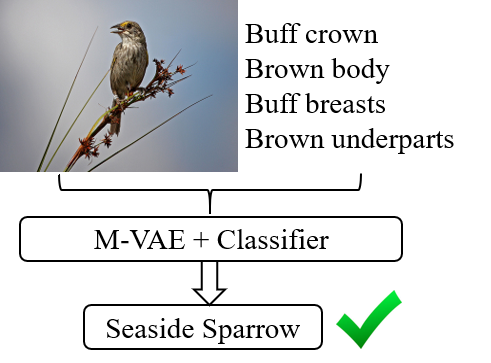}
\end{subfigure}
\caption{Examples showing the classification of images belonging to novel classes from the CUB dataset using M-VAE.}
\label{fig:first_page}
\vspace{-5mm}
\end{figure}

As there is no prior knowledge regarding the novel class, simply recognizing it based on knowledge transfer is inadequate for accurate and robust classification.
To overcome this, Frome et al. \cite{frome2013devise} proposed an extended version of zero-shot learning called Generalized Zero-Shot Learning (GZSL) which used semantic knowledge as class embedding alongside the image features.
This semantic knowledge can be used as prior knowledge towards the task of classifying seen and novel images.
\cite{akata2015label} also show that the use of semantic knowledge can boost the GZSL model performance.
They use harmonic mean (H) of the prediction accuracy over the seen and novel classes to measure the performance \cite{akata2015evaluation}.

\begin{figure*}[t!]
\centering
\includegraphics[width=0.9\linewidth]{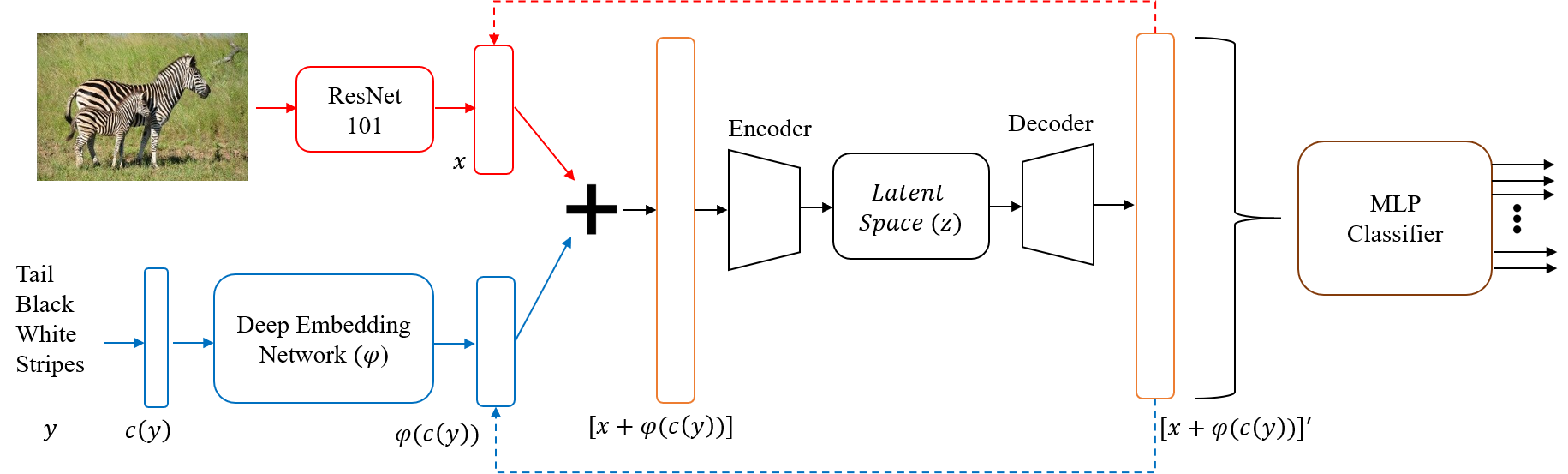}
\caption{Overview of our proposed Multimodal-VAE (M-VAE) architecture. The encoder concatenates the image features and the semantic embedding to the latent space. The decoder reconstructs the original features which are then used to train a classifier.} 
\label{fig:overall}
\vspace{-5mm}
\end{figure*}

GZSL techniques tend to use generative models, such as Variational Auto-Encoder (VAE) and Generative Adversarial Networks (GANs), which reconstruct features for improving the performance.
These feature generation methods generally train a model using either: (1) seen image features and class embeddings \cite{zhu2017unpaired}; (2) novel synthetic features and class embedding \cite{shen2017style}; (3) novel synthetic features and seen image features \cite{schonfeld2019generalized}.
The process of initiation of the novel synthetic features alleviates the data imbalance between the seen and the novel class.
This improves the classification accuracy of the novel classes, thereby improving the harmonic mean (H). 

In recent times, VAEs have shown state-of-the-art performances for reconstructing a feature embedding from the latent space based on cross-modal mapping \cite{kumar2018generalized}.
A VAE-based GZSL architecture transforms different modalities (like image features, class embeddings, semantic knowledge) to a latent space and matches the respective latent space distribution to the original input by reducing the Maximum Mean Discrepancy (MMD) \cite{hubert2017learning}.
Such a cross-modal learning approach can be useful for potential downstream tasks which require multimodal fusion, such as image classification, image captioning, and visual question answering.
Schonfeld et al. \cite{schonfeld2019generalized} proposed a Cross And Distribution Aligned Variational Auto-Encoder (CADA-VAE) for the task of image classification.
Using CADA-VAE, they convert the image features and semantic knowledge into a latent space to overcome the projection domain shift, thereby achieving the then state-of-the-art performance.
Building on top of CADA-VAE, Ma et al. \cite{ma2020variational} proposed a Deep Embedding Variational Auto-Encoder (DE-VAE).
Overcoming the shortcoming of CADA-VAE, the proposed DE-VAE was able to learn the mapping between the image features and the semantic embeddings.
This reduced bias led to a reduction in the difference of the classification performance between the seen and novel classes, thereby improving the harmonic mean (H). 

Even though the DE-VAE model can reduce the bias to a certain extent, the architecture does not much exploit the semantic knowledge.
Exploiting local and global information in the semantic concepts is beneficial to having a robust GZSL model with better performance.
The CADA-VAE model tried to do this using a separate encoder-decoder pair for image features and class embedding, respectively.
Whereas the DE-VAE model used a single encoder to transform both the image features and class embedding to separate latent spaces which are then fed to separate decoders for reconstruction.
Unlike both approaches, we propose to use a single encoder-decoder pair in the VAE for all input modalities combined.
Both CADA-VAE and DE-VAE use cross alignment loss function for reconstruction from separate latent spaces. 
Whereas, in our model, we use a multi-modal loss function for reconstruction which eliminates the need to minimize the distance between the separate latent spaces as done by CADA-VAE and DE-VAE.
Along with this, we use a multimodal (image features and semantic knowledge) loss for the reconstruction of the combined multimodal embedding.
Our hypothesis is that, by using a single encoder-decoder pair along with multimodal loss, we can reduce the misinformation generation towards reconstruction.
Experimental results using our proposed approach on benchmark datasets show performance improvement in the classification of seen and novel samples.
\autoref{fig:first_page} shows examples of successful classification on novel samples from the CUB dataset using our proposed M-VAE model. Following are the major contributions of this work:

\begin{itemize}\vspace{-2mm}
    \item We propose a Multimodal Auto-Encoder (M-VAE) model which can better exploit the global and local semantic knowledge.\vspace{-2mm}
    \item We also propose a multimodal loss which can better reconstruct the image features and semantic knowledge from the latent space.\vspace{-2mm}
\end{itemize}


\vspace{-3mm}
\section{Methodology}

\vspace{-2mm}

\autoref{fig:overall} shows an overview of our proposed Multimodal-VAE (M-VAE) architecture.
Our M-VAE model uses a single encoder-decoder pair for all input modalities combined.
Feature embeddings will be generated individually for each modality, as explained in Section \ref{sec:indi_modality}.
These feature embeddings are combined and used for iterative training and optimization of the VAE to obtain the reconstructed feature embedding.
Section \ref{sec:m_vae} mentions the details of the proposed M-VAE model.
The decoder reconstructs the feature embedding using the multimodal loss, as discussed in Section \ref{sec:m_vae}.
The reconstructed feature embedding, along with the original labels, is then used to train a three-layer MLP (Multi-Layer Perceptron) Classifier.

\begin{table*}[!b]
\centering
\footnotesize
\begin{tabular}{|l|c|c|c|c|c|c|c|c|c|c|c|c|} \hline
\multicolumn{1}{|c|}{Model} & \multicolumn{3}{c|}{CUB (\%)} & \multicolumn{3}{c|}{SUN (\%)} & \multicolumn{3}{c|}{AWA1 (\%)} & \multicolumn{3}{c|}{AWA2 (\%)} \\ \hline 
\multicolumn{1}{|c|}{}      & S      & N      & H      & S      & N      & H      & S       & N      & H      & S       & N      & H      \\ \hline 
CMT \cite{socher2013zero}                         & 49.8   & 7.2    & 12.6   & 21.8   & 8.1    & 11.8   & \textbf{87.6}    & 0.9    & 1.8    & 90.0      & 0.5    & 1.0    \\
SJE \cite{akata2015evaluation}                         & 59.2   & 23.5   & 33.6   & 30.5   & 14.7   & 19.8   & 74.6    & 11.3   & 19.6   & 73.9    & 8.0    & 14.4   \\
ALE \cite{akata2015label}                         & 62.8   & 23.7   & 34.4   & 33.1   & 21.8   & 26.3   & 76.1    & 16.8   & 27.5   & 81.8    & 14.0   & 23.9   \\
LATEM \cite{xian2016latent}                       & 57.3   & 15.2   & 24.0   & 28.8   & 14.7   & 19.5   & 71.7    & 7.3    & 13.3   & 77.3    & 11.5   & 20.0   \\
EZSL \cite{romera2015embarrassingly}                       & 63.8   & 12.6   & 21.0   & 27.9   & 11.0     & 15.8   & 75.6    & 60.6    & 12.1   & 77.8    & 5.9    & 11.0   \\
SYNC \cite{changpinyo2016synthesized}                       & \textbf{70.9}   & 11.5   & 19.8   & 43.3   & 7.9    & 13.4   & 87.3    & 8.9    & 16.2   & \textbf{90.5}    & 10.0   & 18.0   \\
DeViSE \cite{frome2013devise}                     & 53.0   & 23.8   & 32.8   & 27.4   & 16.9   & 20.9   & 68.7    & 13.4   & 22.4   & 74.7    & 17.1   & 27.8   \\
f-CLSWGAN \cite{xian2018feature}                  & 57.7   & 43.7   & 49.7   & 36.6   & 42.6   & 39.4   & 61.4    & 57.9   & 59.6   & 68.9    & 52.1   & 59.4   \\
SE  \cite{kumar2018generalized}                        & 53.3   & 41.5   & 46.7   & 30.5   & 40.9   & 34.9   & 67.8    & 56.3   & 61.5   & 68.1    & 58.3   & 62.8   \\
ReViSE \cite{hubert2017learning}                     & 28.3   & 37.6   & 32.6   & 20.1   & 24.3   & 22.0   & 37.1    & 46.1   & 41.1   & 39.7    & 46.4   & 42.8   \\
CADA-VAE  \cite{schonfeld2019generalized}                  & 53.5   & 51.6   & 52.4   & 35.7   & \textbf{47.2}   & 40.6   & 72.8    & 57.3   & 64.1   & 75.0    & 55.8   & 63.9   \\
DE-VAE \cite{ma2020variational}                     & 52.5   & 56.3   & 54.3   & 36.9   & 45.9   & 40.9   & 76.1    & 59.6   & 66.9   & 78.9    & 58.8   & \textbf{67.4}   \\ \hline
\textbf{M-VAE (ours)}          &  62.9     &  \textbf{57.1}      &   \textbf{59.8}     & \textbf{58.7}       &  42.4      &  \textbf{49.2}     &    78.3     &   \textbf{62.1}     &   \textbf{69.3}    & 72.4        & \textbf{61.3}      &   66.4    \\ \hline
\textbf{Comparison to next best}                     & -8.0   & 0.8   & 5.5   & 15.4  & -4.8   & 8.3   & -9.3    & 1.5   & 2.4   & -18.1   & 2.5   & -1.0   \\ \hline
\end{tabular}
\caption{Experimental results on four benchmark datasets comparing our M-VAE model with state-of-the-art GZSL classification techniques. Best results are marked in bold.}
\label{tab:comparison}
\end{table*}

\vspace{-3mm}
\subsection{Individual Modality Embedding Generation}
\label{sec:indi_modality}
\vspace{-2mm}

In our model, we use two input modalities - image and semantic concepts.
The use of multi-modal input in our model leads to higher accuracy. 
For the image, we extract the image feature embedding $x$ by passing the benchmark dataset images through ResNet-101 model pre-trained on ImageNet dataset.
For the semantic concepts, we use a deep embedding module, as proposed by Ma et al \cite{ma2020variational}, to generate the class embedding  $\varphi(c(y))$. 
We use the deep embedding network to map the semantic space to the image feature space, thereby removing the inherent bias generated by directly using Word2Vec $c(y)$ \cite{schonfeld2019generalized}. 
The deep embedding network comprises of two fully connected linear layers with Rectified Linear Unit (ReLU) activation for each layer.
To fit our M-VAE, the output dimension of the deep embedding network is kept the same as the output dimension of the image features.

\vspace{-3mm}
\subsection{Multimodal-VAE (M-VAE)}
\label{sec:m_vae}
\vspace{-1mm}

Our proposed M-VAE architecture is inspired from Variational Auto-Encoder (VAE) \cite{kingma2013auto}. 
The goal of traditional VAE is to find the true conditional probability distribution of the variables in the latent space, denoted by $p_{\phi}(z|x)$.
Whereas the goal of our M-VAE model is to learn representations within a common space for a combination of multiple data modalities. 
The image feature embedding and semantic embedding, extracted as shown in Section \ref{sec:indi_modality}, are combined to obtain its unified embedding $[x+\varphi(c(y))]$.

CADA-VAE \cite{schonfeld2019generalized} and DE-VAE \cite{ma2020variational} uses multiple encoders / decoders, one for each modality, to map the image features and the semantic information to a latent space.
In our proposed M-VAE model, we use a single encoder to convert $[x+\varphi(c(y))]$ to latent space $(z)$ and a single decoder to reconstruct the embedding $[x+\varphi(c(y))]^{'}$.

\vspace{-3mm}
\subsection{Multimodal Loss}
\label{sec:m_loss}
\vspace{-1mm}

As explained above, the latent space $(z)$ is generated by the encoder $E$ using the combination of different modalities, represented as $[x+\varphi(c(y))]$.
The reconstruction of the combined embedding is done by the decoder $D$ with the associated reconstruction loss given by:

\vspace{-5mm}
\begin{equation}
\label{equ:l_cm}
\mathcal{L} = | [x+\varphi(c(y))]^{(i)} - D(E_i([x+\varphi(c(y))]^{(i)}))|
\end{equation}
\vspace{-2mm}

In our M-VAE model, due to the use of multiple modalities there is a chance of misinformation being generated during reconstruction.
In order to reduce this, we minimize the Wasserstein distance function between the different modalities. The Wasserstein distance between the image feature embedding and semantic embedding is denoted as $W(x,\varphi(c(y)))$. 
Thus, the overall multimodal loss of the M-VAE is represented as:

\vspace{-3mm}
\begin{equation}
\label{equ:overall_loss}
\mathcal{L}_{M-VAE} = \alpha\mathcal{L} + \gamma W(x,\varphi(c(y)))
\end{equation} 
\vspace{-2mm}

where, $\alpha$, and $\gamma$ indicate the weights of the reconstruction loss and the combined Wasserstein distance, respectively.

\begin{table*}[t!]
\centering
\footnotesize
\begin{tabular}{|l|c|c|c|c|c|c|c|c|c|c|c|c|} \hline
\multicolumn{1}{|c|}{Model} & \multicolumn{3}{c|}{CUB (\%)} & \multicolumn{3}{c|}{SUN (\%)} & \multicolumn{3}{c|}{AWA1 (\%)} & \multicolumn{3}{c|}{AWA2 (\%)} \\ \hline 
\multicolumn{1}{|c|}{}      & S      & N      & H      & S      & N      & H      & S       & N      & H      & S       & N      & H      \\ \hline 
Baseline I: w/o Wasserstein distance              & 49.2   & 22.3   & 30.6   & 38.8   & 20.5   & 26.8   & 64.5    & 35.5   & 49.9   & 67.1    & 32.9   & 47.6  \\ 
Baseline II: w/ multiple enc-dec        & 55.2   & 53.2   & 54.0   & 37.3   & 38.8   & 42.2   & 74.4    & 58.9   & 65.7   & 66.6   & 57.4   & 65.5   \\ \hline
\textbf{M-VAE}          &  \textbf{62.9}     &  \textbf{57.1}      &   \textbf{59.85}     & \textbf{58.74}       &  \textbf{42.37}      &  \textbf{49.22}     &    \textbf{78.3}     &   \textbf{62.1}     &   \textbf{69.26}    & \textbf{72.41}        & \textbf{61.3}      &  \textbf{66.39 }   \\ \hline
\end{tabular}
\caption{Results from the ablation study of the proposed M-VAE model.}
\label{tab:ablation}
\vspace{-2mm}
\end{table*}

\vspace{-3mm}
\section{Experiments}
\vspace{-2mm}

\subsection{Benchmark Datasets}
\label{sec:bench_data}
\vspace{-1mm}
The proposed M-VAE model is evaluated on the following four commonly accepted ZSL/GZSL benchmark datasets: 

\vspace{-1mm}
\begin{itemize}
    \item Caltech UCSD Birds 2000-2011 (CUB) \cite{wah2011caltech} dataset contains 312 fine-grained semantics and a total of 200 output classes, of which 150 are used as seen classes and the remaining 50 are used as novel classes. \vspace{-2mm}
    \item SUN Attribute (SUN) \cite{patterson2014sun} dataset contains 102 fine-grained semantics and a total of 717 output classes, of which 645 are used as seen classes and the remaining 72 are used as novel classes. \vspace{-2mm}
    \item  Animals with Attributes 1 (AWA1) \cite{lampert2013attribute} dataset contains 85 fine-grained semantics and a total of 50 output classes, of which 40 are used as seen classes and the remaining 10 are used as novel classes. \vspace{-2mm}
    \item Animals with Attributes 2 (AWA2) \cite{xian2018zero} dataset is a newer version of AWA1 with the same set of classes. \vspace{-2mm}
\end{itemize}

\vspace{-8mm}
\subsection{Implementation Details}
\vspace{-1mm}

Using the pre-trained ResNet-101, we extract the image feature embedding of size 2048.
For the deep embedding network, we use a MLP with one hidden layer of size 1450.
The extracted semantic embedding has size 1200.
As an input to the encoder, we pass the combined embedding of size 3248.
We use 1660 hidden units for both encoder and decoder. 
The latent space size for the M-VAE is fixed to 64.
The M-VAE model is trained for 100 epochs with a batch size of 50 using stochastic gradient descent and Adam optimizer.

\vspace{-3mm}
\subsection{Results}
\label{sec:results}
\vspace{-2mm}
We compare the results of our proposed M-VAE model with the 12 state-of-the-art techniques.
\autoref{tab:comparison} shows the performance results on the four benchmark datasets using accuracy of Seen (S) and Novel (N) classes, as well as the Harmonic mean (H).
We also compare the results from our model to the next best, when it achieves state-of-the-art performance, or compare it to the best result where it does not achieve state-of-the-art results.
We observe that our model achieves state-of-the-art performance for Harmonic mean across all datasets except for AWA2 where the difference is only 1\%.
On an average, our model obtains a 3.8\% improvement across all datasets in comparison to the next best.
We also observe that our model is able to achieve better performance for novel class (N) across the benchmark datasets, except for the SUN dataset.
Our model outperformed the majority of the techniques for the seen class (S) accuracy.

\vspace{-3mm}
\subsection{Ablation Study}
\vspace{-1mm}

\autoref{tab:ablation} shows the results of the ablation study of our proposed M-VAE model using two baselines:

Baseline I: M-VAE reconstructs the combined embedding with the traditional loss, without the Wasserstein distance. As shown in the results, using the proposed multimodal loss (Section \ref{sec:m_loss}) provides a significant performance improvement as compared to the results for the baseline. This confirms that it is beneficial to minimize the distance between the two modalities during the M-VAE training process.

Baseline II: separate modality embeddings and separate encoder-decoder pairs for each modality. From the results, we can validate our claim that we minimize the misinformation generated during reconstruction using a single pair of encoder-decoder as opposed to keeping the embeddings separate and using separate encoder-decoder pairs.

Performing a direct comparison between the results from Baseline I and Baseline II, we can assertively say that minimizing the Wasserstein distance between the modalities has more direct effect on the performance of the model compared to the misinformation generated by using separate encoder-decoder pair of each modality.

\vspace{-3mm}
\subsection{Discussion}
\vspace{-2mm}

As mentioned in Section \ref{sec:results}, our M-VAE achieves significant improvement in the Harmonic mean (H) values for all four datasets, on an average. 
This increase is due to our M-VAE approach of combining multiple modalities into a single embedding before being used for encoder-decoder training.
This approach is intuitively able to optimize the image association with semantic information when reconstructing the embedding.
As shown in the results, we obtain an average reduction of 5\% for the Seen classes (S) compared with the next best state-of-the-art results.
However, we see no reduction in the accuracy for the Novel classes (N).
Also, the difference between the accuracy values for S and N is relatively small when compared to other techniques.
This implies that our M-VAE model is able to adapt to novel classes that may come through during testing.
It is able to maintain a good balance between S and N, by sacrificing the accuracy for the seen classes and not the novel classes, thereby attaining higher harmonic mean.

\vspace{-3mm}
\section{Conclusion}
\vspace{-2mm}

In this work, we introduce a Multimodal Variational Auto-Encoder (M-VAE) for generalized zero-shot learning. 
The model combines the feature embeddings of multiple modalities into a single embedding to be fed to a single encoder-decoder pair for learning the latent space.
To reconstruct the feature embedding, we propose the use of a multimodal loss function that tries to reduce the difference between the individual modalities.
This loss function is a combination of the traditional reconstruction loss and the Wasserstein distance function. 
We performed experimental analysis on four benchmark datasets and compared our approach with 12 state-of-the-art approaches.
The results show that our model outperforms the other approaches for generalized zero-shot learning.
The proposed M-VAE model is able to reduce the misinformation during reconstruction and is also able to exploit the local and global information in the semantic space.
As part of the future work we plan on extending the deep embedding network to incorporate more intricate semantic knowledge and accordingly update the multimodal loss.
This would eventually make our approach more robust and provide improved classification results on the emerging novel classes.

\vspace{-3mm}
\section{Acknowledgement}
\vspace{-2mm}

This project and the preparation of this paper were funded in part by monies provided by CPS Energy through an Agreement with The University of Texas at San Antonio. Copyright© 2020 CPS Energy and The University of Texas at San Antonio. The authors gratefully acknowledge use of the services of Jetstream cloud, funded by NSF award 1445604.

\bibliographystyle{IEEE}
\bibliography{main}

\end{document}